%% file: CBCT_Plan-arxiv.tex
\documentclass[12pt,twoside]{article}
\usepackage[super,sort,comma]{natbib}
\usepackage{fancyhdr}		%Gives headers and footers defined below
\usepackage{xcolor,colortbl,color}
\definecolor{LightCyan}{rgb}{0.88,1,1}
\usepackage{graphicx}
\usepackage[margin=2cm]{geometry}

\usepackage[mathlines]{lineno}

\newcommand*\patchAmsMathEnvironmentForLineno[1]{%
  \expandafter\let\csname old#1\expandafter\endcsname\csname #1\endcsname
  \expandafter\let\csname oldend#1\expandafter\endcsname\csname end#1\endcsname
  \renewenvironment{#1}%
     {\linenomath\csname old#1\endcsname}%
     {\csname oldend#1\endcsname\endlinenomath}}% 
\newcommand*\patchBothAmsMathEnvironmentsForLineno[1]{%
  \patchAmsMathEnvironmentForLineno{#1}%
  \patchAmsMathEnvironmentForLineno{#1*}}%
\AtBeginDocument{%
\patchBothAmsMathEnvironmentsForLineno{equation}%
\patchBothAmsMathEnvironmentsForLineno{align}%
\patchBothAmsMathEnvironmentsForLineno{flalign}%
\patchBothAmsMathEnvironmentsForLineno{alignat}%
\patchBothAmsMathEnvironmentsForLineno{gather}%
\patchBothAmsMathEnvironmentsForLineno{multline}%
}

\usepackage{hyperref}
\hypersetup{ colorlinks,
    citecolor=blue,
    filecolor=blue,
    linkcolor=blue,
    urlcolor=blue
}

\usepackage[all]{hypcap}    %causes link to figures to go to figure, not caption

\usepackage{amssymb,amsmath}

\newcommand{\R}{\mathbb{R}}

\newcommand{\Loss}{\mathrm{Loss}}

\newcommand{\Plan}{\mathbf{Plan}}
\newcommand{\CB}{\mathbf{CB}}

\newcommand{\red}[1]{{#1}}

\begin{document}

\begin{center}
\sf {\Large \bfseries 
Visual enhancement of Cone-beam CT \\ by use of CycleGAN
}

\vspace*{10mm}
S Kida${}^{\dag}$\footnote{Kida and Kaji should be considered joint first authors.},
S Kaji${}^{\ddag}$\footnotemark[1],
K Nawa${}^\dag$, T Imae${}^\dag$, T Nakamoto${}^\dag$, S Ozaki${}^\dag$, T Ohta${}^\dag$, Y Nozawa${}^\dag$ and K Nakagawa${}^\dag$\\
${}^\dag$ Department of Radiology, University of Tokyo Hospital, 7-3-1, Hongo, Bunkyo-ku, Tokyo 113-0033, Japan\\
${}^\ddag$ Institute of Mathematics for Industry, Kyushu University, 744 Motooka, Nishi-ku, Fukuoka 819-0395, Japan / JST PRESTO
\vspace{5mm}\\
\end{center} 

\pagenumbering{roman}
\setcounter{page}{1}
\pagestyle{plain}
Author to whom correspondence should be addressed. email: skaji@imi.kyushu-u.ac.jp \\

\begin{abstract}
\noindent {\bf Purpose:}
Cone-beam computed tomography (CBCT) offers advantages over conventional fan-beam CT
in that it requires a shorter time and less exposure to obtain images.
%CBCT has found a wide variety of applications in patient positioning for image-guided radiation therapy, extracting radiomic information for designing patient-specific treatment, and computing fractional dose distributions for adaptive radiation therapy. 
However, CBCT images suffer from low soft-tissue contrast, noise, and artifacts compared to conventional fan-beam CT images.
Therefore, it is essential to improve the image quality of CBCT. 
\\
{\bf Methods:}
In this paper, we propose a synthetic approach to translate CBCT images with deep neural networks.
Our method requires only unpaired and unaligned CBCT images and planning fan-beam CT (PlanCT) images for training.
The CBCT images and PlanCT images may be obtained from other patients as long as they are acquired with the same scanner settings.
Once trained, three-dimensionally reconstructed CBCT images can be directly translated to high-quality PlanCT-like images.
\\
{\bf Results:}
We demonstrate the effectiveness of our method with images obtained from \red{24} prostate patients,
and provide a statistical and visual comparison.
The image quality of the translated images shows substantial improvement in voxel values, spatial uniformity, and artifact suppression compared to those of the original CBCT.
The anatomical structures of the original CBCT images were also well preserved in the translated images.
\\
{\bf Conclusions:}
%Our method enables more accurate adaptive radiation therapy, and opens up new applications for CBCT that hinge on high-quality images.
\red{Our method produces visually PlanCT-like images from CBCT images
while preserving anatomical structures.}
\end{abstract}

\setlength{\baselineskip}{0.7cm}      %double spacing		

\pagenumbering{arabic}
\setcounter{page}{1}
\pagestyle{fancy}

%\maketitle

%%%%
\input{CBCT_Plan-contents_rev3}

\end{document}

%% file: CBCT_Plan-contents_rev3.tex
\section{Introduction}
Cone-beam computed tomography (CBCT) 
uses a wide, cone-beam of X-rays and offers considerable advantages insofar as the volume information of patients can be obtained more quickly and with lower exposure. 
Compared to conventional fan-beam CT, however, the image quality of CBCT is degraded due to X-ray scattering and truncated projections,
hindering the effective use of CBCT in many potential applications.
It is therefore important to improve the image quality of CBCT.

The most popular use of CBCT in modern radiation therapy is for patient positioning in image-guided radiation therapy  \cite{LetourneauD, GukenbergerM, JaffrayDA, WangX}. CBCT imaging systems can be incorporated into a conventional radiotherapy device, since they are inexpensive and compact. 
Usually, planning fan-beam CT (PlanCT) images with high resolution for treatment planning in radiation therapy are acquired once before the course of treatment.
On the other hand, CBCT images are acquired just before each radiation treatment along the fractional irradiations on the treatment couch to provide up-to-date anatomical information regarding the patients.
Patient setup for radiation therapy is performed by manual or automatic image registration between CBCT and PlanCT using target registration software. 
However, the quality of image registration often depends on the experience and intuition of the operator, because the image quality of CBCT is insufficient for identifying the structure of organs within soft tissue.
%% この辺りは消せと言われるかもしれない。
%% ここは背景ですし、PlanCTからDIRしたものをARTに用いるというのは方向性としては間違ってないので、消さなくていいと思います。今回はそういうレベルまで行ってないことは確かですが。（kida）
Recently, possible applications of CBCT for patient-specific treatment have also been explored. The CBCT images acquired during each treatment can include information on the patient's condition and the response to the treatment. Several methods of quantitative radiomic analysis using CBCT have been reported \cite{FoveX, vanTimmerenJE}. However, the low image quality of CBCT may prevent the accurate extraction of such radiomic information. The effective use of CBCT for adaptive radiation therapy (ART) has also been studied. In addition to CBCT, many approaches to the implementation of ART have been investigated that allow adaptive treatment change based on patient anatomy. Although CBCT images provide up-to-date information regarding patients, they have not been used directly to compute dose distributions for re-planning in ART owing to their low image quality \cite{KimS, StockM}. The conventional approach to dose calculation using CBCT is to transform PlanCT images to CBCT images using deformable image registration (DIR) and then calculate the dose distribution using the transformed PlanCT \cite{KurzC, PaquinD, ParkSB}. Another approach to dose calculation using CBCT is to compute the dose distribution directly with the improved CBCT images \cite{StankovicU}. Improving the CBCT image quality while preserving the anatomical structures of CBCT should be effective for both approaches.

% ここから改良の話
It is essential to improve the image quality of CBCT image for the overall accuracy of radiotherapy.
 Several methods have been applied to remove the scatter photons from CBCT 2D projection images, such as hardware suppression using an anti-scatter grid and software estimation using a scatter deconvolution technique \cite{StankovicU, SisniegaA, SunM}. On the other hand, a different approach has been proposed, which uses PlanCT as prior information to improve the quality of CBCT images. This approach is applicable to direct 3D image reconstruction with sparse sampling \cite{ShiL}, histogram matching \cite{KidarHS}, super-resolution \cite{OyamaA}, and a deep convolutional neural network \cite{KidaS}. Although these do not require access to the raw projection data or a particular scanner manufacturer, they depend on accurate spatial alignment of CBCT and PlanCT volume pairs from the same patients. Misalignments between paired images can lead to errors like edge blurring, deformation, and the disappearance of some anatomical structures in the improved CBCT images. However, it is challenging to achieve sufficiently accurate alignments between CBCT and PlanCT images even for the same patient, because these images are clinically acquired on different days separated by days or weeks. 
 
% NN を使った image processing
 Deep neural networks (DNNs) have been successfully applied to various image processing tasks \cite{MazurowskiMA, higakiT}. In particular, the generative adversarial network (GAN) has been widely used for image-to-image translation \cite{pix2pix}. 
 %Conditional GAN is an approach to learning translation mapping with a conditional setting to capture structural information \cite{pix2pix}. 
 %The objective of a conditional GAN is adversarial loss combined with a voxel-wise loss of the aligned paired images. 
 Nie et al. have applied the idea of combining voxel-wise loss with adversarial loss to translating brain MRI to CT images. As a result, high-quality, less blurry CT images were synthesized \cite{NieD}.
  However, voxel-wise loss also depends on the accuracy of the alignment of paired images, and it is clinically difficult, if not impossible, to acquire aligned paired training images from two different modalities.
To learn the translation mappings in the absence of aligned paired images, a cycle-consistency GAN (CycleGAN) \cite{CycleGAN} has been proposed. 
 Recently, CycleGAN has been used to synthesize CT images from MR images \cite{NieD, WolterinkJM},
 and to synthesize CT images from CBCT images \cite{Liang_2019}.
%In this paper, we adapt a tailored version of CycleGAN to synthesize PlanCT-like images (SynPlanCT) from CBCT images with unpaired and unaligned CBCT and PlanCT training datasets.

%% CycleGAN の説明
 Suppose we have two sets of images from two domains, $X$ and $Y$.
 It should be emphasized that we do not require any prescribed correspondence between the elements of $X$ and $Y$. 
 To translate CBCT and PlanCT images, we need only 
independent images from CBCT and PlanCT, and these may be taken on different days from different patients. We only require that each set is consistent and uniform; the images in each set are acquired with the same scanner setting and contain similar parts of the body.
 CycleGAN learns a mapping $G_{X\to Y}: X\to Y$ and $G_{Y\to X}: Y\to X$, such that the distributions (or characteristics) of images from the output of $G_{X\to Y}$ (and, respectively, of $G_{Y\to X}(Y)$) are indistinguishable from that of $Y$ (and, respectively, $X$) using an adversarial loss. 
  In addition, a cycle-consistency loss
 % $|G_{Y\to X}(G_{X\to Y}(X) - X|$ and $|G_{X\to Y}(G_{Y\to X}(Y) - Y|$ 
 is introduced to force the network to translate the synthesized image back to the original image domain and minimize the difference between the original input image and the reproduced image. 
 %The cycle-consistency is applied to solve the well-known problem in GAN called mode collapse, where all input images are transferred to the same output images. 
 This scheme has intrinsic ambiguity with respect to geometric transformation, which may fail to preserve the anatomical structure in transformed medical images. 

% To mitigate this problem, methods such as shape-consistency loss \cite{ZhangZ} have been introduced to constrain the geometric invariance of synthesized images. However, this requires additional input data (segmentation), which is unavailable in our case insofar as the quality of CBCT images is insufficient to perform reliable segmentation.

 %To our knowledge, however, there has been no research regarding adapting CycleGAN to synthesize PlanCT images from CBCT images. MR images offer higher contrast relative to CT images. Therefore, synthesizing CT from MR images can be performed by reducing the amount of information. On the other hand, compared to CT images, CBCT images have low soft-tissue contrast, noise, and artifacts that spread across on the entire image. In this sense, a non-trivial extension to the network algorithm beyond that of previous works should be constructed in order to synthesize CT from CBCT images.

 In this paper, we propose a synthetic approach to producing PlanCT-like images (SynPlanCT images) from CBCT images. The proposed method relies exclusively on unpaired and unaligned image datasets. 
 The proposed approach is based on CycleGAN with modifications tailored for this specific task. 
 The network structure and the loss functions were determined through trial and error while observing the image quality of the SynPlanCT images. 
 In addition to the image quality, we paid a particular attention to structure preservation.
 We devised a few tricks to enforce boundary-preservation (typically, that of bone and air),
which clinicians rely on for image registration.
 The image quality of SynPlanCT images was quantitatively evaluated in terms of their voxel values and spatial uniformity. We analyzed the extent to which they preserved the anatomical structures through a comparison with the original CBCT images, and the corresponding PlanCT images aligned to the CBCT images by the deformable image registration.
 %PlanCT was not used for analysis of anatomical preservation.}
 The robustness of our method was confirmed by investigating SynPlanCT images from five network models trained with the same structure and hyper-parameter settings. 
% To the best of our knowledge, this is the first study that attempts to translate CBCT images to PlanCT images with a DNN using only unpaired and unaligned training data.
 
 %PatchGAN was adopted to preserve high frequency structures. 
 %Evaluation results show that the image qualities of SynPlanCT images have shown a substantial improvement on voxel values, spatial uniformity and suppression of artifacts compared to those of original CBCT and the anatomical structures of CBCT were well preserved in SynPlanCT.
 %The reproducibility of our method was demonstrated by investigating SynPlanCT images from five network models trained with the same hyper-parameter setting. To the best of our knowledge, this is the first study that attempts to translate CBCT images to PlanCT images with deep neural networks using only unpaired training data.

The remainder of this paper is organized as follows.
In \S \ref{sec:methods}, we explain how we collected and processed image data to train our neural networks,
and we provide details regarding the neural networks based on CycleGAN. We also discuss the evaluation of our method,
which is based on comparing the ROI from the images. 
In \S \ref{sec:results}, we present the output of our method and some statistics related to its evaluation.
In \S \ref{sec:discussion}, we focus on the evaluation, specifically the robustness of our method and its improvement in image quality.
In \S \ref{sec:conclusion}, we summarize our findings.

%%%%%%%%%%%%%%%
\section{Materials and Methods}\label{sec:methods}
\subsection{Data acquisition and image processing}\label{sec:dataset}
 In this study, we used CBCT and PlanCT images from prostate cancer patients who underwent stereotactic radiotherapy with an Elekta Synergy linear accelerator (Elekta AB, Stockholm, Sweden).
The PlanCT images were acquired on a 16-row multidetector helical CT scanner with a tube voltage of 120 kV, a tube current of 350 mA, a gantry rotation time of 0.5 s, a matrix size of 512 by 512 on the axial plane with a pixel size of 1.074 mm by 1.074 mm and a slice thickness of 1 mm. The PlanCT images were reconstructed by an analytical reconstruction method in the form of filtered backprojection (FBP).
 %We used between 170 and 210 slices per patient. 
 The CBCT images were acquired during the course of the treatment using a kV on-board imager (XVI) with a tube voltage 120 kV, a tube current 40 mA/frame, and an exposure time 40 ms/frame. For each scan, in total, 360 projections were acquired in a full-scan ($360^\circ$ gantry rotation). The CBCT reconstructions were performed by the analytical reconstruction method in the form of FBP using projection images of one imaging panel shifted to 11.5 cm to encompass the target. These were output to match the resolution and the slice thickness of the PlanCT images automatically using XVI.
For 16 patients, the PlanCT images were acquired only once, about two weeks before the course of treatment. 
These images were used for training.
%Among the images from the 20 patients, those from 4 patients (referred to hereafter as (i)--(iv)) were kept for evaluation, and those from the remaining 16 patients were used to train the network.
 In addition, there were 4 patients from whom PlanCT images was acquired just after the acquisition of the CBCT images (they are referred to hereafter as patient (i)--(iv)). 
 In those CBCT and PlanCT images, the anatomical structures were relatively close.
 We transformed the PlanCT images to align with the CBCT images by a deformable image registration using RayStation (v4.6, RaySearch Laboratories).
    We used these deformed PlanCT images (hereafter, they are referred to as DefPlanCT images)
    for evaluation. Ideally, the result of our method should be evaluated
    using the CBCT and PlanCT images, which are acquired at the same time,
    and thus make a perfectly matched pair. 
    Such a pair is virtually impossible to obtain, and we suppose that 
    our DefPlanCT is a reasonable alternative.
 We used a total of 2795 CBCT images and 2795 PlanCT images for training, although these numbers need not be same.

Preprocessing to obtain training CBCT and PlanCT images was performed as follows. To prevent any adverse impacts from non-anatomical structures (treatment couch and other objects outside the body) on a CBCT-to-PlanCT registration and as a model training procedure, binary masks were created to separate the pelvic region from the non-anatomical regions.
These masks were generated by applying the Otsu auto-thresholding method \cite{otsu} on each CBCT and PlanCT image. The voxel values outside the mask region were entirely replaced with a Hounsfield Unit (HU) of -1000. Then, the masked PlanCT volume data for each patient were three-dimensionally pre-aligned to each of the masked CBCT images by rigid registration using an open-source software called Elastix \cite{KleinS}. 
Though CycleGAN does not require exactly aligned paired images of CBCT and PlanCT, three-dimensional (3D) pre-alignment was performed such that the bodies of all patients were included in a cropped calculation area around the center with a size of $480\times 384$ pixels for efficient calculation. 

%%%%
\subsection{Image synthesis with deep neural network}\label{sec:networks}
We regard a (volumetric) CT image in any modality as an array $\R^{h\times w \times d}$,
where $h\times w$ is the slice dimension and $d$ is the number of slices.
Image synthesis (or conversion) can be considered as a mapping 
$\R^{h_1\times w_1 \times d_1}\to \R^{h_2\times w_2 \times d_2}$,
which takes an image of size $h_1\times w_1$ and $d_1$ channels/slices (a CBCT image, in our particular case) 
and outputs another of size $h_2\times w_2$ and $d_2$ channels/slices (a PlanCT image, in our case).
We assume that $h_1=h_2=h$ and $w_1=w_2=w$ by resizing the images if necessary.
A popular way to construct such a mapping is to use a DNN
that learns (i.e., is trained) to construct mappings from a large set of data.
The problem of synthesizing a PlanCT image that corresponds to a given CBCT image is highly ill-posed.
That is, there is no unique ``best'' way to find the solution.
Thus, we define what constitutes bad mappings in a quantitative way in terms of a loss function,
whereby the DNN finds an approximate mapping that minimizes the loss function. 

Due to computational limitations, we assume that
the conversion mapping is uniform along the third dimension such that it can be approximated by
a mapping $\R^{h\times w}\to \R^{h\times w}$.
That is, our mapping converts CT images slice-by-slice along the $z$-axis.

Our strategy is largely based on CycleGAN \cite{CycleGAN}, 
which simultaneously constructs four mappings:
(i) a generator $G_{C\to P}: \R^{h\times w}\to \R^{h\times w}$,
     which takes a CBCT image and outputs a synthesized PlanCT image,
(ii) a generator $G_{P\to C}: \R^{h\times w}\to \R^{h\times w}$,
    which takes a PlanCT image and outputs a synthesized CBCT image,
(iii) a discriminator $D_{P}: \R^{h\times w}\to \R^{h'\times w'}$, 
    which distinguishes synthesized PlanCT images from real ones
(iv) and a discriminator $D_{C}: \R^{h\times w}\to \R^{h'\times w'}$, 
    which distinguishes synthesized CBCT images from real ones.
The discriminator $D_{P}$ looks at (overlapping) patches in a whole slice 
and determines whether each patch is likely to be one from a real image.
Ideally, $D_{P}$ would output the all-zero array $\mathbf{0}$ 
for synthesized PlanCT images and the all-one array $\mathbf{1}$ for real PlanCT images.
The discriminator $D_{C}$ works similarly for real and synthesized CBCT images.
We approximate these mappings by training a DNN for each mapping from 
a set $\Plan$ of PlanCT images and a set $\CB$ of CBCT images.
%The details of the dataset are given in \S \ref{sec:dataset}.
To this end, we formulate the desired properties of the mappings in terms of loss functions.
Generators and discriminators have different sets of goals, and hence, different loss functions.
Let $||v||_p$ denote the $L^p$-norm of a vector $v$.
Our loss function for the discriminators is based on LSGAN \cite{LSGAN}.
%% weighted version
%Each discriminator $D$ outputs two channels $D(x)_1$ and $D(x)_2$, where $D(x)_1$ the usual discriminating output and $D(x)_2$ is the weighting:
%\begin{align*}
%\Loss_{D} = & \lambda_D \sum_{x\in \CB} ||
%(\mathrm{tanh}(D_P(G_{C\to P}(x))_2)+ \mathbf{1})\odot (D_P(G_{C\to P}(x))_1-\mathbf{0})\odot (D_P(G_{C\to P}(x))_1-\mathbf{0})||_1 \\
%& \qquad + ||(\mathrm{tanh}(D_C(x)_2+ \mathbf{1}) \odot (D_C(x)_1-\mathbf{1})\odot (D_C(x)_1-\mathbf{1})||_1 \\
%   & +\lambda_D\sum_{y\in \Plan}   ||\mathrm{tanh}(D_C(G_{P\to C}(y))_2+ \mathbf{1}) \odot (D_C(G_{P\to C}(y))_1-\mathbf{0})
%   \odot (D_C(G_{P\to C}(y))_1-\mathbf{0})||_1 \\
%   & \qquad + ||(\mathrm{tanh}(D_P(y)_2+ \mathbf{1}) \odot (D_P(y)_1-\mathbf{1})\odot (D_P(y)_1-\mathbf{1}) ||_1  \\
%  &  + \lambda_{reg} (\sum_{x\in \CB}   ||D_C(x)_2||^2_2 + \sum_{y\in \Plan}||D_P(y)_2||^2_2) ,
%\end{align*}
%where $\odot$ denotes the element-wise product.
\begin{align*}
\Loss_{D} = & \lambda_D \sum_{x\in \CB} \left( ||D_P(G_{C\to P}(x))-\mathbf{0}||_2 + ||D_C(x)-\mathbf{1}||_2 \right)\\
   & +\lambda_D\sum_{y\in \Plan} \left( ||D_C(G_{P\to C}(y))-\mathbf{0}||_2 + ||D_P(y)-\mathbf{1}||_2 \right) .
\end{align*}
The loss function forces $D_P$ (and, respectively, $D_C$) to be trained to distinguish the distributions of 
real PlanCT (and, respectively, CBCT) images from synthesized ones.
%The idea of having weighting is to tell discriminators where to focus on.
%For example, a patch of size greater than the receptive field of the discriminator (in our case, $73\times 73$)
%which contains only air cannot be distinguished in real and synthesized images.
%Without the weighting, discriminators would put much effort on such discrimination since it fails by a 50 percent chance.
%The learned weighting is supposed to direct the effort of discriminators towards meaningful regions.

Our loss function for generators $G_{C\to P}$ and $G_{P\to C}$ consists of several terms:
\begin{itemize}
    \item $\Loss_{cycle}=\Loss_{cycleA}+\Loss_{cycleB}$, where $\Loss_{cycleA}= \sum_{x\in \CB}||x - G_{P\to C}(G_{C\to P}(x))||_1$, $\Loss_{cycleB}=\sum_{y\in \Plan}||y - G_{C\to P}(G_{P\to C}(y))||_1$.
    This is the cycle-consistency loss which ensures that $G_{C\to P}$ and $G_{P\to C}$ are inverse mappings with respect to each other.
    \item $\Loss_{adv}=\sum_{x\in \CB}||D_P(G_{C\to P}(x))-\mathbf{1}||_2 + \sum_{y\in \Plan}||D_C(G_{P\to C}(y))-\mathbf{1}||_2$:
    This encourages the generators to fool the discriminators by synthesizing realistic images.
%    \item \begin{align*}
%    \Loss_{adv}=& \lambda_D \sum_{x\in \CB} ||
%(\mathrm{tanh}(D_P(G_{C\to P}(x))_2)+ \mathbf{1})\odot (D_P(G_{C\to P}(x))_1-\mathbf{1})\odot (D_P(G_{C\to P}(x))_1-\mathbf{1})||_1 \\
%   & +\lambda_D\sum_{y\in \Plan}   ||\mathrm{tanh}(D_C(G_{P\to C}(y))_2+ \mathbf{1}) \odot (D_C(G_{P\to C}(y))_1-\mathbf{1})
%   \odot (D_C(G_{P\to C}(y))_1-\mathbf{1})||_1 \\
%   \end{align*}
%    This is the adversarial loss which encourages the generators to fool the discriminators by synthesizing realistic images.
    \item $\Loss_{tv}=\sum_{x\in \CB}||\mathrm{grad} G_{C\to P}(x)||_1$,
    where $\mathrm{grad}$ is the image gradient.
    This is the total variation regularization which encourages the generator to produce spatially uniform images.
    Since even the real images in $\CB$ are noisy, we apply this loss only to $G_{C\to P}(x)$ which are the SynPlanCT images.
    \item $\Loss_{air}=\sum_{x\in \CB}||\psi(G_{C\to P}(x))-\psi(x)||_1+\sum_{y\in \Plan}||\psi(G_{P\to C}(y))-\psi(y)||_1$, where 
    $\psi(z)=\begin{cases} z & (z< C) \\ 0 & (z\ge C)\end{cases}$,
    where $C$ is a constant equivalent to $-465$ HU.
    This enforces the generators not to alter air regions having low values less than $-465$ HU
     to preserve the air--body boundary.
%    \item $\Loss_{perc}=\sum_{x\in \CB}||VGG(G_{C\to P}(x))-VGG(x)||_2+\sum_{y\in \Plan}||VGG(G_{P\to C}(y))-VGG(y)||_2$,
%     where $VGG$ is a pretrained VGG16 network truncated at the \textbf{conv2\_2} layer. This is the perceptual loss \cite{artistic} based on the DNN pretrained with natural images in ImageNet.
%    This encourages the generators to produce perceptually similar images for human eyes to the input images 
%    by preserving the gradient and other common image features learned from natural images.
    \item $\Loss_{grad}= \sum_{x\in \CB}(||\partial_1(x - G_{C\to P}(x))||_2+||\partial_2(x - G_{C\to P}(x))||_2) 
    + \sum_{y\in \Plan}(||(\partial_1(y - G_{P\to C}(y))||_2+\partial_2(y - G_{P\to C}(y))||_2)$,
    where $(\partial_1,\partial_2)$ is Sobel's approximated gradient operator~\cite{Sobel}.
    This encourages structural preservation before and after conversion by trying to keep the edges in the image. 
    \item $\Loss_{idem}= \sum_{x\in \CB}||G_{C\to P}(x) - G_{C\to P}(G_{C\to P}(x))||_1 
    + \sum_{y\in \Plan}||G_{P\to C}(y) - G_{P\to C}(G_{P\to C}(y))||_1$:
    This ensures that $G_{C\to P}^2=G_{C\to P}$ and $G_{P\to C}^2=G_{P\to C}$ (i.e., that they are idempotent).
\end{itemize}
The terms $\Loss_{cycle}$ and $\Loss_{idem}$ pertain to mathematical requirements regarding what the conversion mappings should satisfy,
and they help to increase the stability during training.
The terms $\Loss_{grad}$ and $\Loss_{adv}$ are competing demands as the former stipulates that the generator preserves 
the structure of the input image while the latter encourages altering the input image to look more like a real one.
In the end, they should find a good balance.
%We could extend $\Loss_{air}$ to cover other special regions such as bones, 
%but we noticed that some CBCT artifact has high HU values similar to those of bones.
Note that $\Loss_{cycle}$ indirectly encourages structure preservation by enforcing translation maps to be 
one-to-one, but it does not guarantee structure preservation as is seen in Fig. \ref{fig:fail}.
Incorporating $\Loss_{grad}$ and $\Loss_{air}$ suppress the type of large deformation in translation.
It is also important to note that CycleGAN (and generally, GAN) was developed to have a wide variety of output, 
while in our medical setting, it is important to have stable outputs.
The newly incorporated loss terms all contribute to increase the regularity of the optimisation problem, and thus, 
the stability.

We combine these terms by taking the weighted sum with respect to the hyper-parameters $\lambda$:
\[
 \Loss_G = \lambda_{cycle} \Loss_{cycle} + \lambda_{adv} \Loss_{adv} + \lambda_{grad} \Loss_{grad} + \lambda_{idem} \Loss_{idem}
 + \lambda_{air} \Loss_{air}+ \lambda_{tv} \Loss_{tv}
%  + \lambda_{perc} \Loss_{perc} 
\]
and use stochastic gradient descent to find a (local) minimizer of the combined loss function.
%In practice, we can add more terms as regularizers to exploit prior knowledge from the image domains.
%See our implementation \cite{codes} for details.
%Our contributions in the design of the loss function are 
%(i) the novel definition of the discriminator loss $\Loss_{D}$ using patch weighting, and 
%(ii) the region specific preservation loss $\Loss_{air}$ to promote preservation of anatomical structures.

%\begin{figure}[ht]
%\center
%     \includegraphics[width=0.9\linewidth]{fig1.pdf}
%     \caption{Overview of the CycleGAN model with a forward cycle and a backward cycles.}\label{fig:cycle}
%\end{figure}

%%%%%
\subsubsection*{Technical details}\label{sec:details}
For efficiency, the intensity of CT images was clipped to $[-500,200]$ HU,
and scaled to $[-1,1]$. That is, the pixels with an HU of less than -500 were all mapped to -1,
and those with an HU of higher than 200 were all mapped to 1.
When fed into the network, the images were cropped randomly around the center
to a size of $480\times 384$ pixels. 
This size was selected such that the body was always contained well inside the edges in order to avoid the 
boundary effects of convolutional neural networks in general.

For the generators $G_*$, we used an encoder--decoder network consisting of
(i) One convolution layer with a $7\times 7$ kernel with stride $1$,
(ii) followed by three down convolution layers with a $3\times 3$ kernel with stride $2$ and channels $32,64$ and $128$,
(iii) followed by $9$ residual blocks with a $3\times 3$ kernel with stride $1$,
(iv) followed by three up-sampling layers each consisting of an unpooling with stride $2$ followed by a residual block with a $3\times 3$ kernel with stride $1$,
(v) followed by a convolution layer with a $7\times 7$ kernel with stride $1$.
Except for the last layer, we used the rectified linear unit for activation, and instance normalization \cite{InstanceNormalization} for normalization.
The last layer was equipped with the hyperbolic tangent as activation and without normalization.
Layers (i),(ii) and (iv),(v) have skip connections \cite{U-net} to transfer the structure of the input image directly.
To increase the stability, we added Gaussian noise of $\sigma=0.05$ to the latent variable, which 
is just after the 4th residual block.

For the discriminators $D_*$, we used a typical down convolution network consisting of
(i) A convolution layer with a $4\times 4$ kernel with stride $1$ and channels $32$,
(ii) followed by three down convolution layers with a $4\times 4$ kernel with stride $2$ and channels $64,128$ and $256$,
(iii) followed by a convolution layer with a $4\times 4$ kernel with stride $1$ and channels $256$,
(iv) followed by a convolution layer with a $4\times 4$ kernel with stride $1$ and channels $1$.
We used a leaky rectified linear unit with slope $0.2$ for activation, and instance normalization for normalization in all but the last layer.
We did not apply any normalization to the last layer.
The receptive field of the network was $73\times 73$ and each pixel in the output revealed the evaluation for the patch of this size
in the input image.
We also experimented with different network configurations by modifying the kernel size of each layer and the total number of layers,
and thus with different receptive field sizes, and observed that the above configuration performs well.
 
In both the generators and the discriminators,
it is important to use instance normalization without an affine transform; with an affine transform, we observed that
the output images tended to have low voxel values in certain regions.
The padding value for the first layer was selected to match the intensity of the air ($-1$, in our case).

We chose the hyper-parameters empirically as follows:
\[
\lambda_{cycle}=10.0, \lambda_{adv}=1.0,  \lambda_{grad}=0.1, \lambda_{tv}=0.01, \lambda_{air}=1.0, \lambda_{idem}=1.0, \lambda_D=1.0.
\]
All networks were trained from scratch at a learning rate of $10^{-4}$, with the Adam optimizer with a batch size of $1$. 
We kept the same learning rate for the first 25 epochs, and linearly decayed the rate to zero over the next 25 epochs.

Note that for generators, minimizing $\Loss_{adv}$ means maximizing $\Loss_{D}$,
and for discriminators, minimizing $\Loss_{D}$ means maximizing $\Loss_{adv}$.
Generators and discriminators are adversarial, trying to beat each other.
If the competition is nearly even, they both improve in a steady manner.
Therefore, when training GANs, their strength must be balanced.
There are several ways to help to maintain this balance:
(i) tuning the weightings $\lambda_{adv}$ and $\lambda_{D}$,
(ii) using different loss functions (e.g., Wasserstein GAN with a gradient penalty \cite{WGAN}),
(iii) scheduling updates for the generator and discriminator (e.g., updating the generator only once for every five updates of the discriminator),
(iv) tuning the degrees of freedom of the networks for the generators and discriminators, and
and (v) defining good stopping criteria.
Although it was unfeasible to try all the possibilities,
we observed that the last one had the largest effect. We thus adjusted the number of channels in each network.
Although Wasserstein GANs are believed to be very helpful, in our experiments we could not find a good configuration for our particular problem.
Therefore, we decided to stick to the current loss function.
Similarly, it was appropriate to set $\lambda_{adv}=\lambda_{D}$ and to update the generators and discriminators in equal frequency.
The point at which learning stopped was also important. 
We noticed that if we trained too many epochs, the preservation of anatomical structures began to deteriorate.
We watched the learning progress by visually inspecting the output, and decided to train 50 epochs for this study.
However, this number should be dependent on the dataset. 
Establishing judicious stopping criteria remains for future research.

%The weights of the neural networks were randomly initialized using He's method \cite{He}.
%In order to mitigate instability in the initialization, we set the first 200 iterations as a warmup period.
%During the warmup, we set $\lambda_{adv}=0$,
% and added an extra term
%$\Loss_{id} = 
%\sum_{x\in \CB}||x - G_{C\to P}(x)||_2 + \sum_{y\in \Plan}||y - G_{P\to C}(y)||_2$
%to $\Loss_G$. 
%This means that the generators first learn the identity mapping,
%and then proceed to learn the desired conversion. The loss function for the discriminator remains unmodified during the warmup period.
%Note that during the initial stage, discriminators output nonsense, so it does not make sense to use $\Loss_{adv}$
%to train the generators.

%%%
%\subsection{Environment and timing}
We conducted experiments with a personal computer equipped with a single GPU (Nvidia 1080Ti) 
and a CPU (Intel Core i7-6950X) with  132 GB memory, running Ubuntu 18.04.1 LTS.
We implemented our algorithm with Python 3.6.6 and Chainer 5.0.0 \cite{chainer}.
Our codes are available on Github\footnote{\url{https://github.com/ shizuo-kaji/UnpairedImageTranslation}}
%For the CycleGAN code, we based our implementation on \url{https://github.com/~naoto0804/~chainer-cyclegan}.}.
The training required approximately 20 minutes per epoch (2795 iterations),
and 13 hours for 50 epochs.
The conversion required about 6 seconds for one volume of data (32 slices per second).

%Progressive GAN,

\subsection{Evaluation method}
\begin{figure}[!htb]
 \center
    \includegraphics[width=0.8\linewidth]{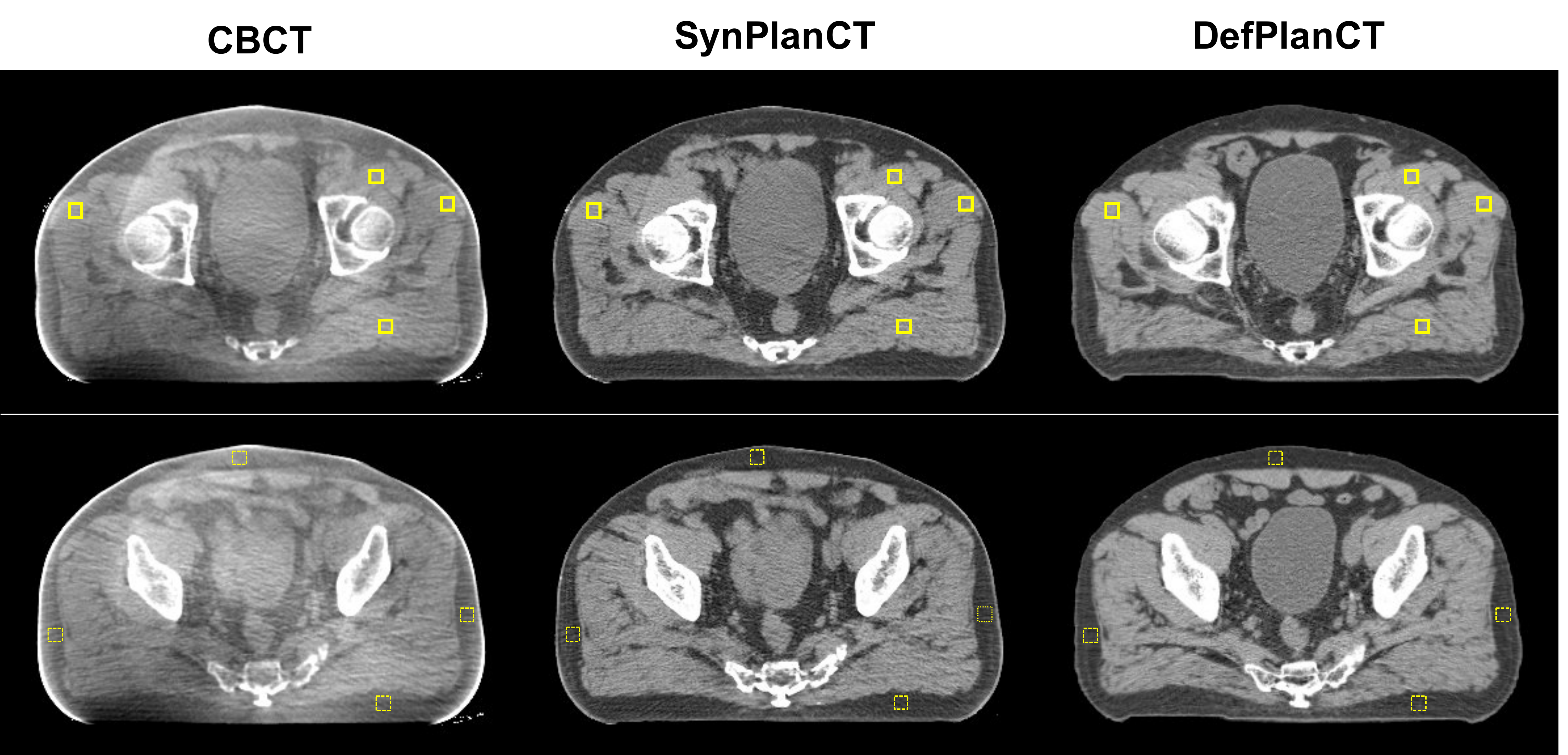}
     \caption{Example of evaluating the ROI distribution: (Upper left) Four muscle ROIs (solid square) placed on a selected slice of CBCT for test patient (iv), (Upper middle) corresponding ROIs placed on a corresponding slice of SynPlanCT, and (upper right) corresponding ROIs placed on a corresponding slice of DefPlanCT. (Lower left) Four fat ROIs (dotted square) placed on another slice of the CBCT, (lower middle) corresponding ROIs placed on a corresponding slice of the SynPlanCT, and (lower right) corresponding ROIs placed on a corresponding slice of the DefPlanCT; The display window range was set to (-400, 0) HU for CBCT and (-200, 200) HU for SynPlanCT and DefPlanCT.}\label{fig:ROI}
\end{figure}

The CBCT and PlanCT image pairs from 4 prostate cancer patients (i)--(iv) were used for the quantitative evaluation. Each PlanCT image was acquired just after each corresponding CBCT image was acquired, and the PlanCT images were aligned to the CBCT images by DIR (see \S \ref{sec:dataset}).
The resulting SynPlanCT images were compared to the original CBCT and DefPlanCT images using the HU for statistic and visual inspections of the regions of interest (ROIs) from four different types of tissues (viz., prostate, bladder, muscle, and fat). For a quantitative evaluation of the muscle and fat regions, four slices were selected, one slice from each of the four patients. 
Four square ROIs with $10 \times 10$ pixels were positioned in the regions of the same soft tissue area (fat or muscle) in distant locations on a selected slice (Figure \ref{fig:ROI}). 
Four slices at 3 mm intervals per patient were selected for the prostate region. Four slices at 6 mm intervals per patient were selected for the bladder region. One square ROI with $10 \times 10$ pixels was positioned in the regions of the same soft tissue area (prostate or bladder) on a selected slice. These ROIs were positioned on the corresponding slices and locations of the CBCT, SynPlanCT and DefPlanCT images. That is, 16 ROIs were positioned for each soft tissue to evaluate the distribution of the HU of each ROI. 
All the ROIs were selected from the region where DIR worked relatively well.

To determine the robustness of our method, we used images from patients (i)--(iv). For each evaluation slice, we produced five SynPlanCT images from five network models trained with the same structure and hyper-parameters.
    
%%%%%%%%%%%%%%%%%%%%%
\section{Results}\label{sec:results}
\subsection{Comparison of Image qualities and Preservation of anatomical structures}
\begin{figure}[!htb]\center
     \includegraphics[width=0.9\linewidth]{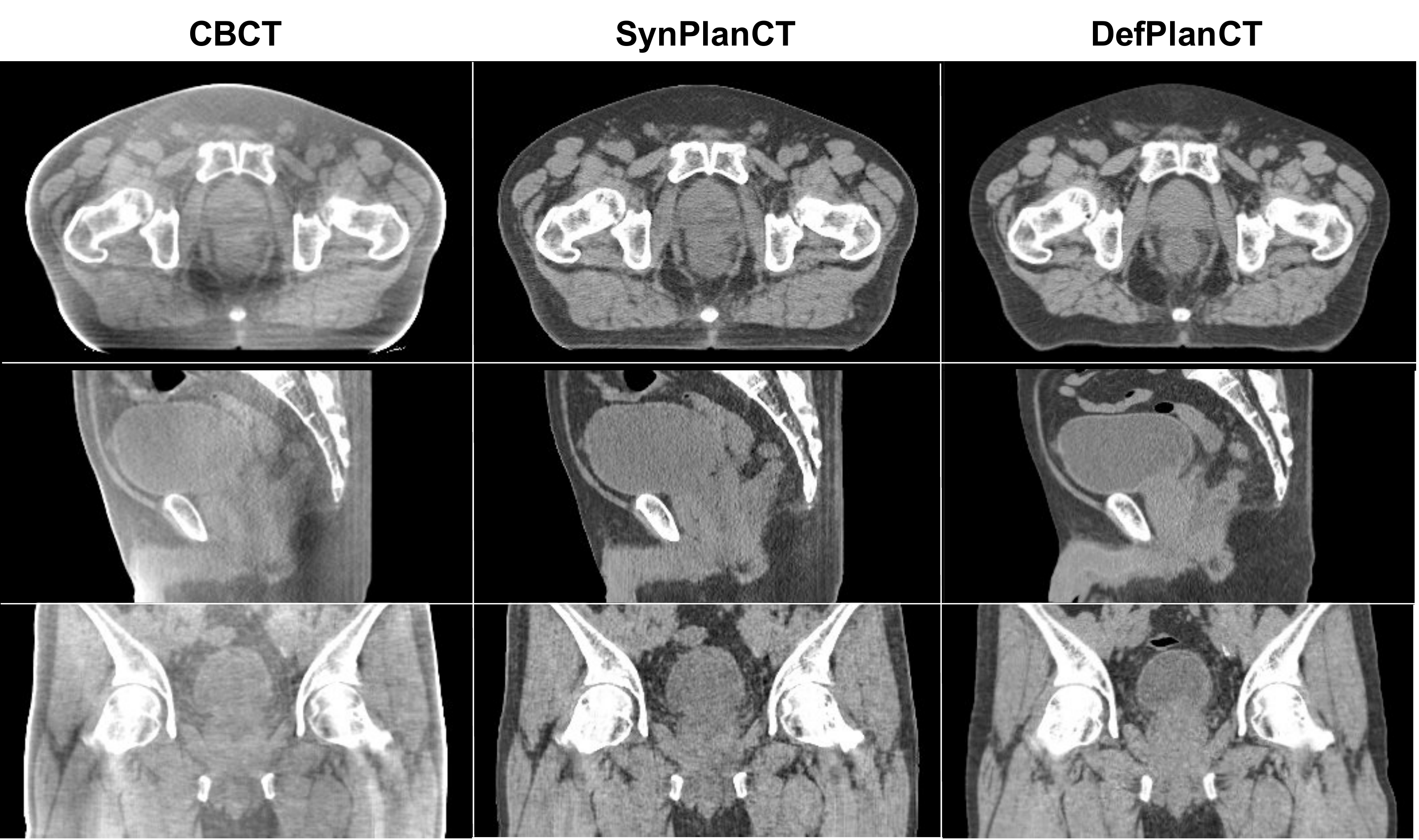}
     \caption{Comparison of the image quality among CBCT, SynPlanCT, and PlanCT \red{of test patient (ii)}. For each patient, the images in the top, middle, and bottom row are axial, coronal, and sagittal views, respectively. The images on the left, middle, and right are CBCT, SynPlanCT and PlanCT, respectively. The display window range was set to (-400, 0) HU for CBCT and (-200, 200) HU for SynPlanCT and PlanCT.}\label{fig:slices}
\end{figure}

The axial, sagittal, and coronal slices of CBCT, SynPlanCT, and PlanCT for two representative cases (\red{test patient (ii)}) are shown in Figure \ref{fig:slices}. The image quality of SynPlanCTs for all three planes show substantial improvement in terms of voxel values, spatial uniformity, artifact suppression, compared to those of the original CBCT. 

\begin{figure}[!htb]\center
     \includegraphics[width=0.9\linewidth]{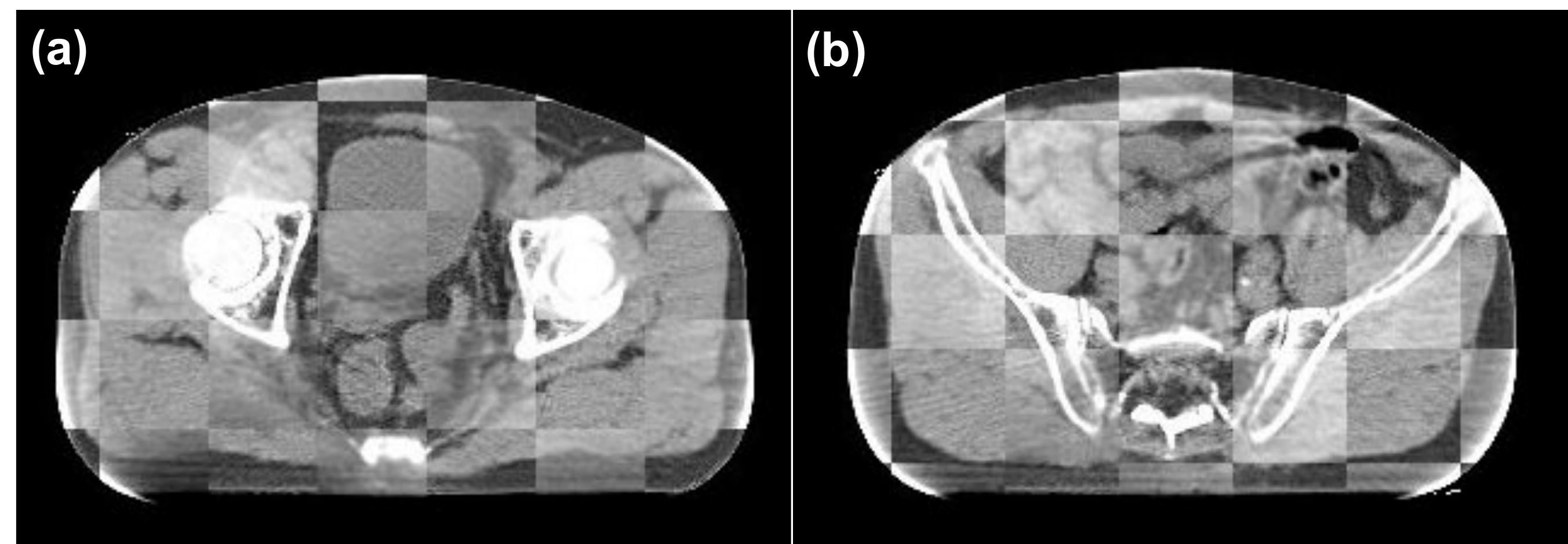}
     \caption{Checkerboard images of a CBCT image and SynPlanCT image: (a) bladder and rectum slice, and (b) small intestine slice. The display window range was set to (-400, 0) HU for CBCT and (-200, 200) HU for SynPlanCT.}\label{fig:checker}
\end{figure}

 As shown in Figure \ref{fig:checker}, the edge sharpness and structure were preserved in not only the large bulky tissues, such as the rectum and bladder, but also in small isolated structures, such as the small intestine and intestinal gas. High-frequency artifacts such as streaks and rings could not be completely removed, but they were considerably suppressed. In this study, despite learning with only axial slices, owing to computational limitations, there was no outstanding problem in the continuity of the structure and voxel values in the other two planes (coronal and sagittal), as shown in Figure \ref{fig:slices}.

\vspace{5mm}

\vspace{5mm}

\vspace{5mm}
\begin{figure}[!htb]\center
%\begin{minipage}{0.45\textwidth}
     \includegraphics[width=0.9\linewidth]{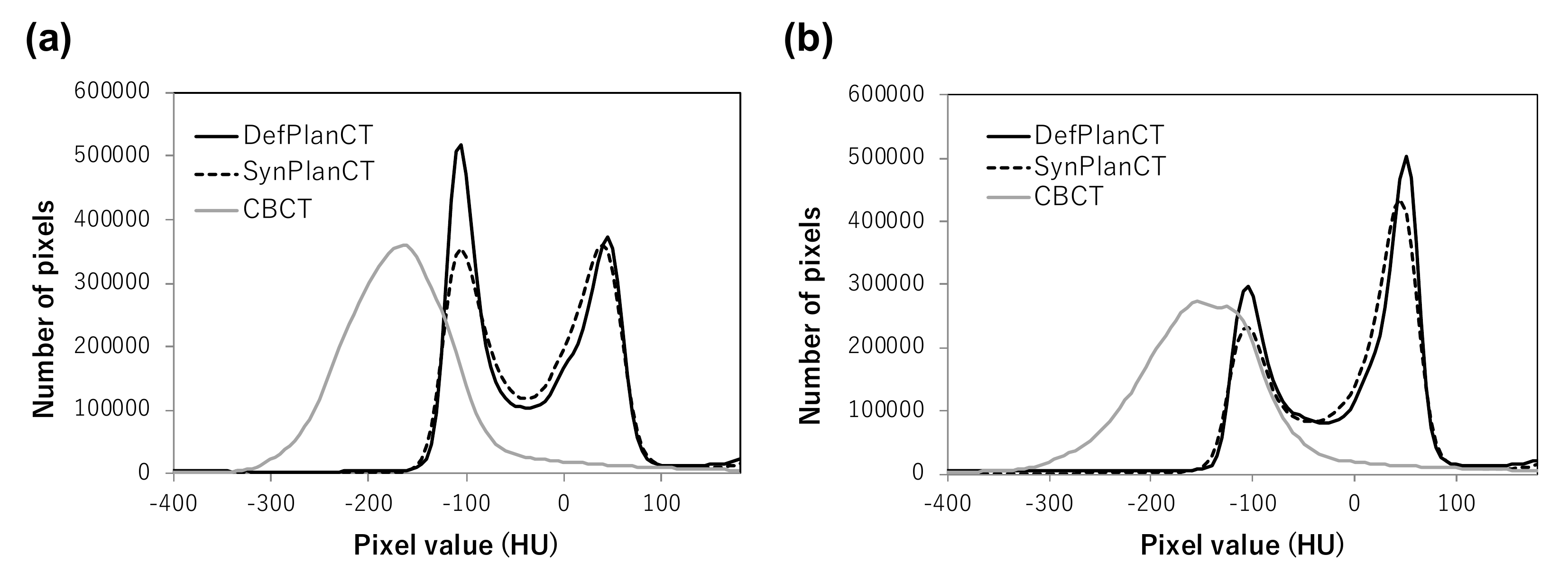}
     \caption{Image histograms for the 3D volume of CBCT (grey line), corresponding volume of SynPlanCT (dotted black line) and corresponding volume of DefPlanCT (solid black line) for representative two patients. (a) patient (ii) and (b) patient (iii)}.\label{fig:hist}
%\end{minipage}
\end{figure}

\subsection{Quantitative evaluation} 
\begin{figure}[!htb]
\vspace{1cm}
\center
     \includegraphics[width=0.45\linewidth]{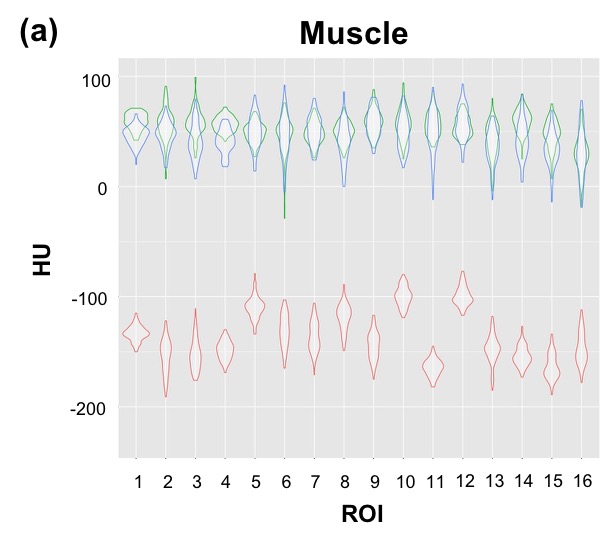}
     \includegraphics[width=0.45\linewidth]{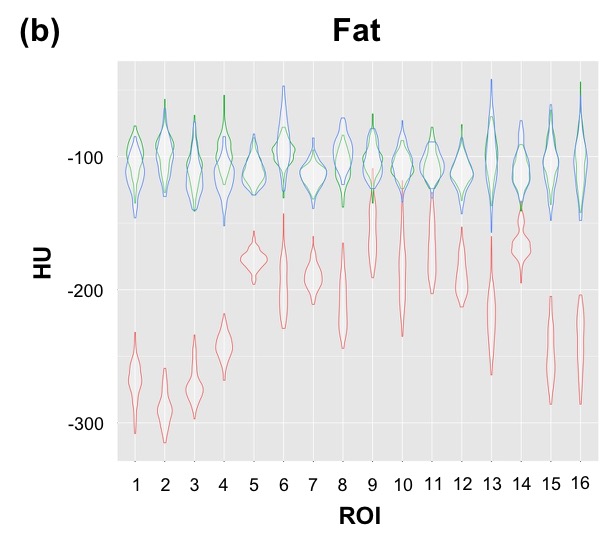}\\
   \includegraphics[width=0.45\linewidth]{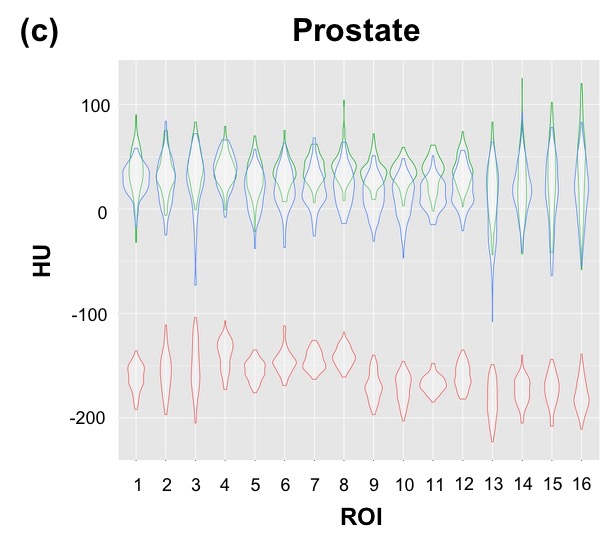}
  \includegraphics[width=0.45\linewidth]{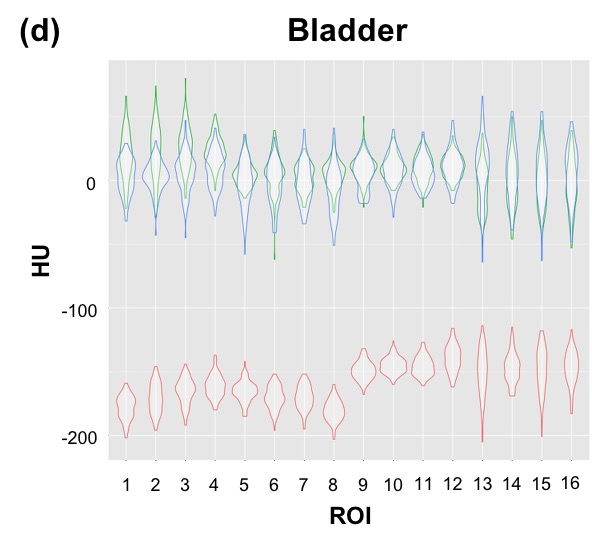}
\caption{Comparison of the distribution of the HU values in (a) muscle, (b) fat, (c) prostate, and (d) bladder ROIs among CBCT (red), SynPlanCT (blue), and DefPlanCT (green). ROI(1-4), ROI(5-8), ROI(9-12), and ROI(13-16) are selected for test patients (i), (ii), (iii), and (iv), respectively. The width indicates the ratio of the pixels with a specific HU.
     }\label{fig:comparison}
%     \caption{Comparison of the distribution of the average HU of each ROI positioned in the four evaluation tissues (upper-left, muscle; upper-right, fat; lower-left, bladder; lower-right, prostate) between SynPlanCT and PlanCT.}\label{fig:comparison}
\end{figure}

The histograms of the 3D volume of CBCT, SynPlanCT, and DefPlanCT images for two representative cases (test patients (ii) and (iii)) are shown in Figure \ref{fig:hist}. The histograms of SynPlanCT images are much closer to those of the PlanCT images, with better differentiation between the fat and muscle, compared to those of the CBCT images.
The mean and standard deviation of the HU of the 16 ROIs positioned in the four evaluation tissues in DefPlanCT, SynPlanCT, and CBCT are summarized in Table \ref{tab1}.
For each tissue type, 16 ROIs were selected from four patients.
%, where the average HU of an ROI refers to the average voxel value of $10\times 10$ pixels in the ROI. 
The difference in the average HU for the four evaluation tissues between SynPlanCT and DefPlanCT (muscle: 7HU, fat: 2HU, prostate: 14HU, bladder: 4HU) were substantially suppressed compared to those between the original CBCT and DefPlanCT (muscle: 190HU, fat: 110HU, prostate: 194HU, bladder: 166HU).
%Figure \ref{fig:muscle} shows a comparison of the original CBCT, SynPlanCT, and PlanCT in the distribution of the average HU of each muscle ROI as an example. The variance of the average HU of each muscle ROI for SynPlanCT is much closer to that for PlanCT compared to that for CBCT. 
Figure \ref{fig:comparison} shows comparisons among CBCT, SynPlanCT, and DefPlanCT in the distribution of the HU of each ROI positioned in the four evaluation tissues. 
    The width indicates the ratio of the pixels with a specific HU.
    We observe that the HU distributions of SynPlanCT are much closer to those of DefPlanCT than those of CBCT for all the ROIs positioned in the four evaluation tissues.
%The variance of HU of ROI positioned in the four evaluation tissues for SynPlanCT was larger than those for PlanCT. There were a few ROIs with specifically high HU in fat and low HU in the prostate. 

\vspace{5mm}
\begin{table}[!htb]
\center
\begin{tabular}{|>{\columncolor[gray]{0.8}}c|cccc|}
  \hline
%\rowcolor{LightCyan}
  \hspace{1cm} & & DefPlanCT (HU) & SynPlanCT (HU) & CBCT (HU) \\
  \hline
  Muscle & mean & 52 & 45 & -138 \\
         & sd   & 14 & 17 & 23 \\ \hline
	Fat  & mean & -104 & -106 & -214 \\
	  & sd & 13 & 16 & 43 \\ \hline
	  Prostate & mean & 33 & 19  & -161 \\
	  & sd & 23 & 22  & 19  \\ \hline
	  Bladder & mean & 8 & 4 & -158 \\
	  & sd & 18  & 18  & 17  \\ \hline        
  \end{tabular}
     \caption{Mean and standard deviations of the HU values of the ROIs in the four evaluation tissues in DefPlanCT, SynPlanCT, and CBCT.}\label{tab1}
\end{table}

Assessment of the quality of an image in the absence of a reference image
is called as non-reference image quality assessment (NR IQA) \cite{IQA}.
The NR IQA metrics are defined for natural images but not for medical purposes.
In fact, we tried some metrics including
Natural Image Quality Evaluator (NIQE) and Blind/Referenceless Image Spatial Quality Evaluator (BRISQUE) only to find that they were highly dependent on the ROIs
and were thus not robust.
Instead, we propose a simple metric to assess the resolution and contrast of the images,
which we found to be very stable:
\[
 \mathrm{SelfSSIM}(img)=\mathrm{SSIM}(img,blur),
\]
where $\mathrm{SSIM}$ is the Structural Similarity \cite{SSIM}
and $blur$ is the Gaussian blur of $img$ with $\sigma=3$.
If an image $img$ has good spatial resolution and contrast, 
it should have a large difference if blurred.
Therefore, the lower the $\mathrm{SelfSSIM}(img)$ is, the more the quality of $img$ is.
Table \ref{tab:SSIM} shows the mean and standard deviation of
$\mathrm{SelfSSIM}$ for 30 ROIs from each of CBCT, DefPlanCT, and SynPlanCT of size $120 \times 120$ located at the center.
The HU values are scaled to [0,255] and we used an implementation
of SSIM in OpenCV 4.1.0 \cite{OpenCV}.

\begin{table}[!htb]
\center
\begin{tabular}{|>{\columncolor[gray]{0.8}}c|ccc|}
  \hline
%\rowcolor{LightCyan}
  \hspace{1cm} & DefPlanCT (HU) & SynPlanCT (HU) & CBCT (HU) \\
  \hline
  mean & 0.575 & 0.688 & 0.799 \\ 
  sd & 0.0080 & 0.0095 & 0.0159 \\
  \hline
  \end{tabular}
     \caption{Mean and standard deviation of $\mathrm{SelfSSIM}$ of the ROIs in the four evaluation tissues in DefPlanCT, SynPlanCT, and CBCT.}\label{tab:SSIM}
\end{table}

\begin{figure}[!ht]\center
     \includegraphics[width=0.7\linewidth]{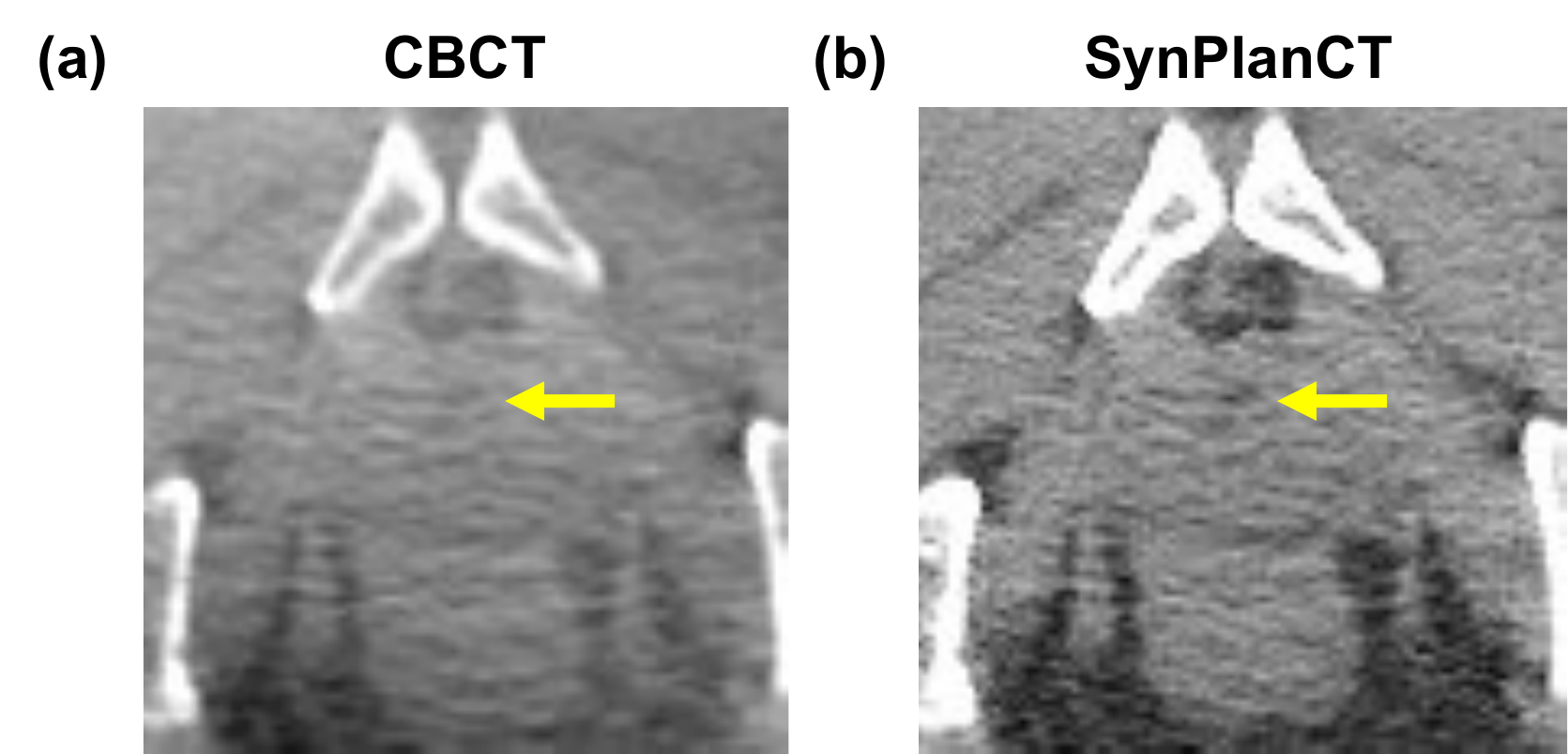}
     \caption{Example of regions with particularly low HU in a prostate: (a) Region with lower HU than the surrounding prostate region is indicated by the arrow on the selected slice of CBCT for test patient (iv). (b) Region with lower HU than the surrounding prostate region is indicated by the arrow on the corresponding slice of SynPlanCT of the same patient. The indicated regions were partially included in ROI 13 of Figure \ref{fig:comparison}(c).}\label{fig:difficult}
\end{figure}

There was an ROI with specifically low HU in the prostate for the SynPlanCT images, as shown in the ROI 13 of Figure \ref{fig:comparison}(c). Figure \ref{fig:difficult} compares the CBCT and SynPlanCT images synthesized by applying a network model (m1) to test patient (iv). We found that the voxel values of the region indicated by the arrow on the SynPlanCT images were lower than the surrounding prostate region (Figure \ref{fig:difficult}(b)), and the region was included in ROI 13 partially. Low signal artifact was observed originally in the prostate region of the CBCT image, as indicated by the arrow in Figure \ref{fig:difficult}(a), and it seemed that the low signal artifact was emphasized with a higher contrast on SynPlanCT (Figure \ref{fig:difficult}(b)). However, it is difficult to determine whether this low signal is an artifact or a tissue feature. It may be important in terms of structural preservation to preserve subtle contrasts, when it is difficult to demarcate artifacts from tissues.

\begin{figure}[!ht]\center
%\begin{minipage}{0.45\textwidth}
     \includegraphics[width=0.8\linewidth]{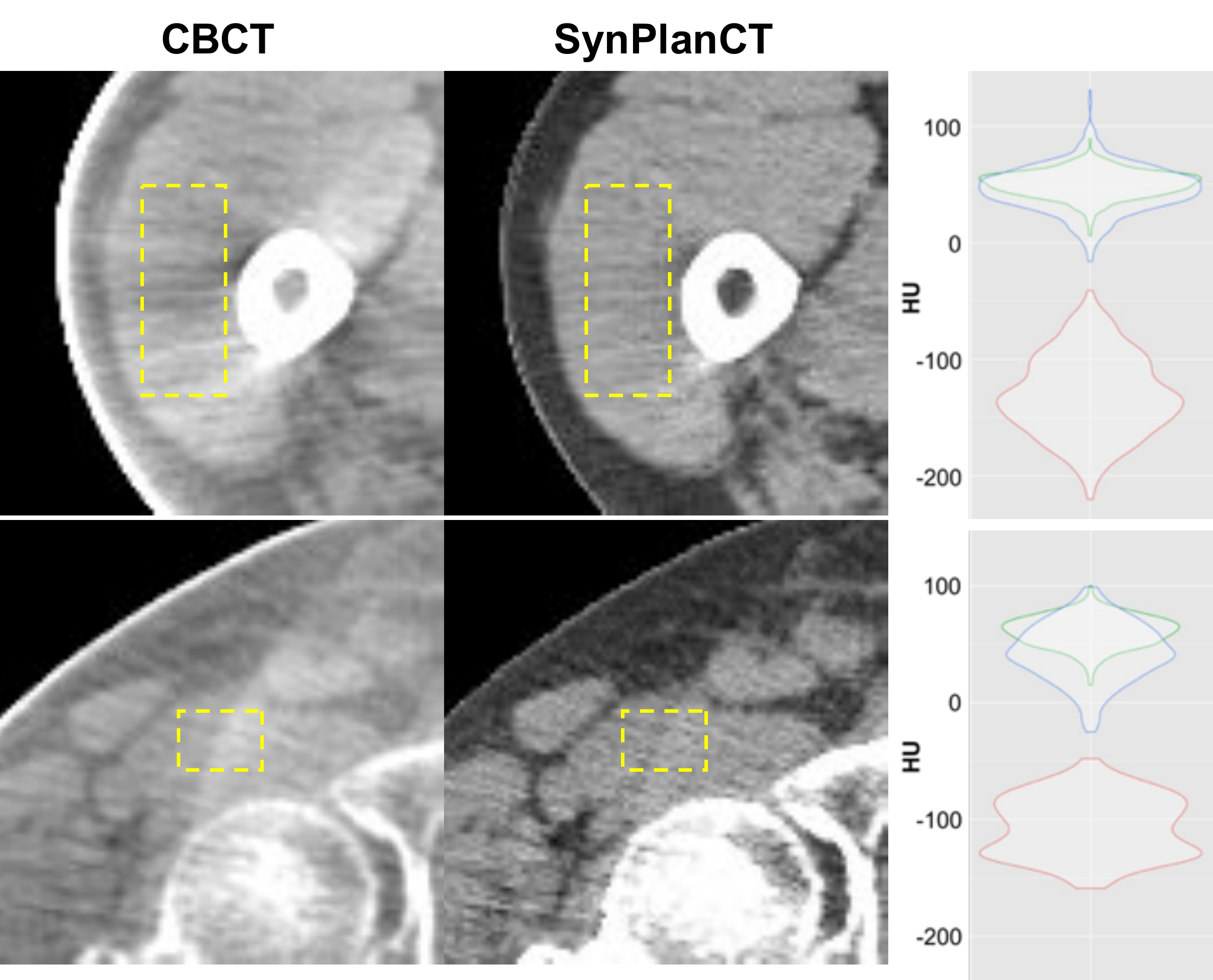}
     \caption{Comparison of the image quality of the artifact region for CBCT and SynPlanCT. (Upper-left) Evaluation ROI ($20\times 50$ pixels) was placed on the fat region of a selected slice of CBCT for test patient (ii), (upper-middle) corresponding ROI placed on a corresponding slice of SynPlanCT of the same patient, and (Upper-right) the distributions of the HU of the selected evaluation ROIs for CBCT (red), SynPlanCT (blue), and DefPlanCT (green). (Lower-left) Evaluation ROI ($20\times 15$ pixels) was placed on the muscle region of the selected slice of CBCT for test patient (iv), (lower-middle) corresponding ROI placed on the corresponding slice of SynPlanCT of the same patient and (lower-right) the distributions of the HU of the selected evaluation ROIs for CBCT (red), SynPlanCT (blue), and DefPlanCT (green). The width indicates the ratio of the pixels with a specific HU.
     }\label{fig:artifact}
%\end{minipage}
\end{figure}

Figure \ref{fig:artifact} compares the image quality of the artifact region between CBCT and SynPlanCT and the distributions of the HU of the selected evaluation ROIs for the representative two slices. Evaluation ROI ($20\times 50$ pixels) was placed on the fat region of the selected slice of CBCT for test patient (ii) and the corresponding ROI was placed on the corresponding slice of SynPlanCT of the same patient (Figure \ref{fig:artifact}(upper row)). 
Another evaluation ROI ($20\times 15$ pixels) was placed on the muscle region of the selected slice of CBCT for test patient (iv) and the corresponding ROI was placed on the corresponding slice of SynPlanCT of the same patient (Figure\ref{fig:artifact}(lower row)).
The streak and ring artifacts were substantially suppressed visually for both the cases. For the distributions of HU, large differences of HU between the tissues and artifacts were observed as different peaks for CBCT; however, such differences of HU were hardly observed for SynPlanCT and DefPlanCT in both the cases. 

%Figure \ref{fig:models} compares the distributions of the HU of one ROI positioned in each of the four evaluation tissues of each of the five SynPlanCT images, synthesized using the five network models trained with the same structure and hyper-parameters (m1--m5) and those of PlanCT. Note that the SynPlanCT images used in Figure \ref{fig:slices}--\ref{fig:comparison} and Table \ref{tab1} were synthesized by applying the network model (m1). The average HU differed by 10 HU or less for all of the four evaluation tissues in the five SynPlanCT images. 

%\begin{figure}[!ht]\center
%\begin{minipage}{0.45\textwidth}
%     \includegraphics[width=0.5\linewidth]{fig4.pdf}
%     \caption{Comparison of the distribution of average HU in muscle ROIs among CBCT, SynPlanCT, and PlanCT.}\label{fig:muscle}
%\end{minipage}
%\end{figure}

\vspace{5mm}

\vspace{5mm}

%\begin{figure}[!htb]\center
%%\begin{minipage}{0.45\textwidth}
%     \includegraphics[width=0.8\linewidth]{fig6.pdf}
%     \caption{Comparison of the distribution of the HU of one ROI positioned in each of} the four evaluation tissues for each of the five SynPlanCT images from five network models trained with the same structure and hyper-parameters (m1--m5) and those of PlanCT for comparison.
%     The width indicates the ratio of the pixels with a specific HU.
%     \label{fig:models}
%%\end{minipage}
%\end{figure}

\begin{figure}[!htb]\center
     \includegraphics[height=2.5cm]{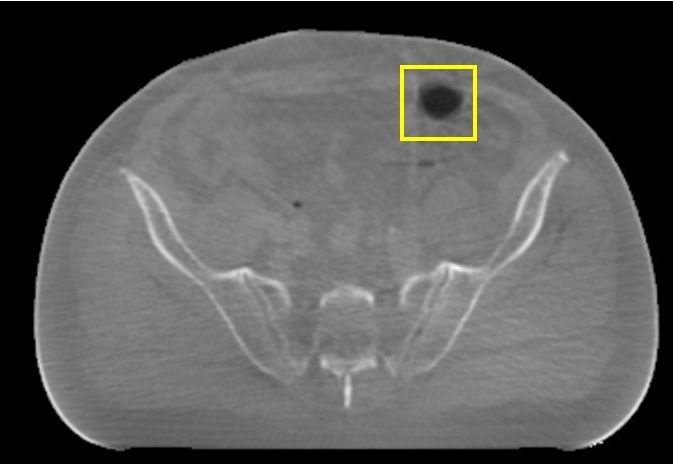}
     \includegraphics[height=3.0cm]{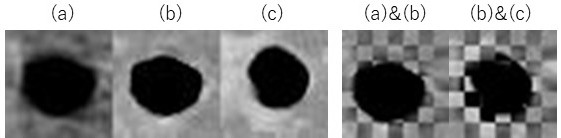}
     \caption{
\red{Comparison of a ROI containing air. From left to right, ROI position in CBCT,
     (a) Original CBCT, (b) SynPlanCT with $\lambda_{air}=1.0$ and $\lambda_{grad}=0.1$,
     (c) SynPlanCT with $\lambda_{air}=0$ and $\lambda_{grad}=0$,
     checkerboard overlay of (a) and (b), and checkerboard overlay of (a) and (c).
     We can see the shape of the air regions in (a) and (b) match very well.
     }}
     \label{fig:air}
\end{figure}

%%%%%%%%%%%%%%%%%%%%%%%%%%%
\section{Discussion}\label{sec:discussion}
%% CycleGAN の話。特に alingment 
 We developed a CycleGAN-based method to synthesize PlanCT-like images from the CBCT images.
 In contrast to the previous work using PlanCT on the same patient as prior information, the proposed method does not, in principle, require accurate spatially aligned CBCT and PlanCT volume pairs from the same patients. 
 In this study, models were trained with CBCT and PlanCT images of the same patients, and these images were rigidly registered to include bodies of all patients in a cropped calculation area around the center, in the size of $480\times 384$ pixels for efficient calculation.
 During the training process of CycleGAN, the discriminators were provided randomly selected unpaired and unaligned images, rather than aligned paired images. Therefore, registration between CBCT and PlanCT should not affect the image quality of synthesized PlanCT. 
% (Note: removed PatchGAN)
 %In addition, by virtue of PatchGAN, the discriminators look only at a small patch and try to determine whether each patch in an image is real or fake, canceling the effects of rigid registration. 
 %In fact, when using non-aligned CBCT and PlanCT volume pairs from the same patient, which were shifted randomly to each other by 50 pixels or more, the image quality improved and the structure was better preserved with similar quality to those of the aligned ones. This result shows that the proposed method does not require aligned volume datasets. Furthermore, in order to evaluate learning with completely unpaired and unaligned images,
% 学習データの必要量
The amount of training data does affect the quality of synthesized PlanCT.
To see this, the 16 patients for training were divided into two groups. 
Then, the network model was trained with the CBCT images of eight patients and the PlanCT images of the other eight patients. 
This model showed similar improvement in the voxel values, spatial uniformity, and artifact suppression on the SynPlanCT images, 
but the structures were incorrectly synthesized in some slices. 
The failure to preserve the structures may be caused by over-learning due to the reduction by half of the learning data. 

%% Evaluation の話
In evaluating a medical application for a DNN, the following three points of view are important.
(i) Accuracy: the output of the method is close to the ground truth, which is often clinically unavailable.
(ii) Precision: the method is robust and produces stable outputs. In general, DNN-based methods inevitably involve randomness when initializing the weights of the networks and during the learning process. Hence, even with the same parameters and the same dataset, their outcome can never be the same.
(iii) Generalization: the method works with different datasets acquired in different ways.
%% Accuracy
To evaluate the accuracy of our method ideally, we compare the SynPlanCT images with the PlanCT images acquired simultaneously with the CBCT images, although these pairs are obviously unavailable. 
Hence, in this study, we assessed the accuracy of the output SynPlanCT by visually inspecting 
The preservation of te anatomical structures compared with CBCT, and the image quality compared with PlanCT.

%% 構造の保存について（Fig.3 & 5 & air）  
Preserving anatomical structures is crucial for image-improvement methods using unpaired and unaligned CBCT and PlanCT datasets. In a previous work using PlanCT as prior information, high-frequency artifacts such as streaks, blurred edges, deformations, and missing anatomical structures were left as problems to be solved \cite{ShiL, KidarHS, OyamaA, KidaS}. In this study, the image quality of CBCT improved while suppressing high-frequency artifacts and preserving the anatomical structures of CBCT. Designing the size of the receptive field of the discriminator is important for preserving the structure and concurrently converting the voxel values. When the receptive field is set too large, learning with a DNN is influenced by the structure and placement of the organs of individual patients. That is, it requires overlearning. On the other hand, when the receptive field is too small, the local structural pattern cannot be detected, and only the voxel values are converted, ignoring the structure. The $73\times 73$ receptive fields of the discriminator used in this study can detect typical local structural patterns commonly found in all patients, enabling the conversion of voxel values while preserving the structure (Figure \ref{fig:checker}). 
    Another important factor which contributes to 
    the preservation the anatomical structures is our design of the loss functions
    for the neural networks (see \S \ref{sec:details}). 
    In particular, if we set $\lambda_{air}=0$ and $\lambda_{grad}=0$,
    some anatomical structures of CBCT were altered in SynPlanCT.
    Figure \ref{fig:air} shows an example of a ROI from CBCT and
    SynPlanCT produced by a mode trained with $\lambda_{air}=1.0,  \lambda_{grad}=0.1$ and one with $\lambda_{air}=0,  \lambda_{grad}=0$.
    They show exactly the same coordinates but the shape of the air region was altered in SynPlanCT with $\lambda_{air}=0,  \lambda_{grad}=0$.
    
%% 3軸での学習と構造の連続性について （Fig.3） 
 Moreover, despite learning with only axial slices, due to computational limitations, there was no outstanding problem in the continuity of the structure and voxel values in the coronal and the sagittal planes (Figure \ref{fig:slices}). Such continuity along the other two planes is evidence that the anatomical structure was correctly preserved within the axial images. In future work, we will investigate how learning directly with volume data can affect the continuity of voxel values and the preservation of anatomical structures.
 
%% 画素値の定量的評価について（Table.1）
 The image quality of SynPlanCT showed substantial improvement in terms of voxel values, spatial uniformity, and artifact suppression compared to those of the original CBCT. 
 %As shown in Table 1, the differences in the maximum and minimum average HUs of muscle and fat ROIs on SynPlanCT (44 and 83 HU) were much smaller than those of the original CBCT (268 and 218 HU), while those of prostate and bladder ROIs on SynPlanCT (45 and 34 HU) and CBCT (42 and 41 HU) were relatively close. 
 The variation of distributions of HU among fat ROIs was larger than that of other tissues (muscle, prostate, bladder) as shown in Table \ref{tab1} and Figure \ref{fig:comparison}. This result means that various artifacts, such as rings streaks, and shading have a stronger effect on fat, spreading closer to body surface, than in the muscle, prostate, and bladder, located relatively near the center of the body, and indeed the artifacts were effectively suppressed by the proposed method. In this study, voxel values of CBCT and PlanCT images were  clipped to [-500, 200] for efficient calculation. Thus, the voxel values of bone, which are more than 200 HU, were not evaluated quantitatively. In future work, we will expand the range of voxel values used for learning and quantitatively evaluate the improvement in voxel values of bone.
 
\begin{figure}[ht]
\center
     \includegraphics[width=0.7\linewidth]{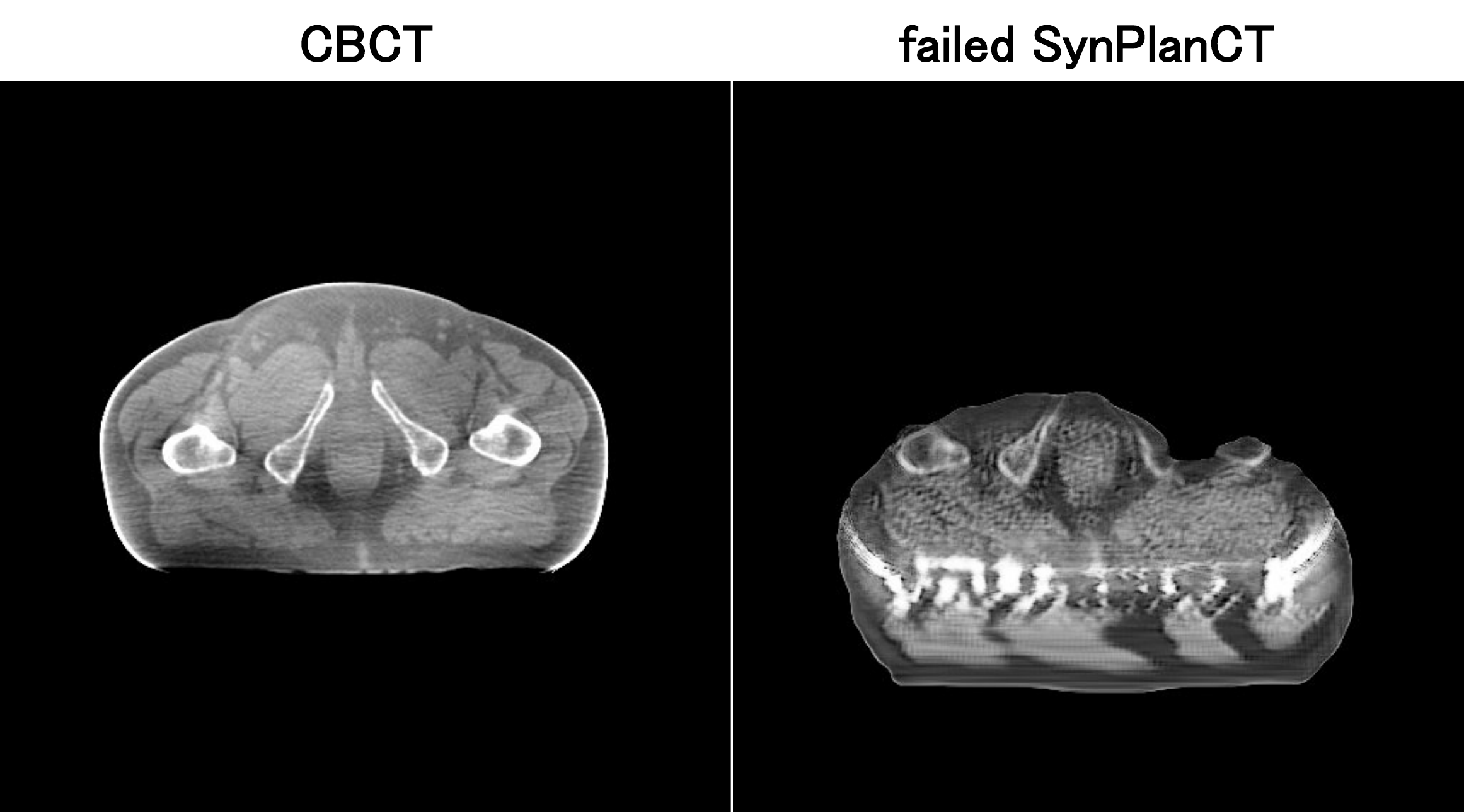}
     \caption{Example of a failed SynPlanCT image from some network model: (left) original CBCT image, and (right) failed SynPlanCT image.}\label{fig:fail}
\end{figure}

%% Precision
In order to assess the precision of our method,
we compared five models trained with the same structure and hyper-parameters, but trained with different random initial weights and stochastic gradient descent. We observed acceptable fluctuation of less than 10 HU. %(Figure \ref{fig:models}).
Note that we did observe on one occasion that models trained with the same hyper-parameters produced totally different images.
Although it is customary in medical applications of DNNs to present results from only the best models,
we here provide an extreme example in which the generators
(trained with $\lambda_{grad}=0$ and $\lambda_{air}=0$)
completely failed to learn a meaningful mapping 
and produced totally distorted images (see Figure \ref{fig:fail}).
In our method, this type of failure was mitigated by introducing additional loss terms $\lambda_{grad}, \lambda_{tv}$, and $\lambda_{air}$ as discussed in \S \ref{sec:details}, 
which helped to increase the overall robustness.
Even after proper training, it is impossible to ensure that the DNN will always produce feasible outputs.
Thus, it is more important to detect and alert the user when a failure may have occurred.
We can use the cycle consistency $\Loss_{cycleA}$ (see \S \ref{sec:networks}) as a measure of the soundness of our networks' output
 in some cases.
Some kind of failures (e.g., see Figure \ref{fig:fail}) can be detected as a high value of $\Loss_{cycleA}$, namely, when this is more than three times that when training was successful.
Recall that $\Loss_{cycleA}$ is a summary of the
difference in the voxel values of the CBCT image $x$ and 
the cyclically translated image $G_{P\to C}(G_{C\to P}(x))$.

\begin{figure}[ht]
\center
     \includegraphics[width=0.9\linewidth]{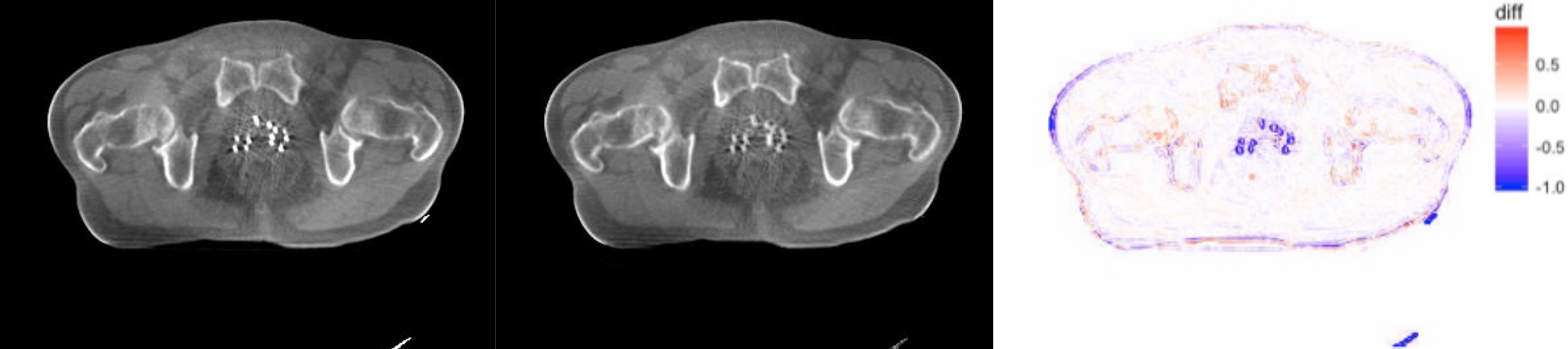}
     \caption{Example of cycle difference: (left) original CBCT image containing metal seeds, (middle) cyclically translated image,
     and (right) their difference in the gradient.}
     \label{fig:cyclediff}
\end{figure}

We can look directly at a so-called difference image $G_{P\to C}(G_{C\to P}(x))-x$
to locate where the violation of cycle consistency occurred. 
To visualize the violation, it is more suitable to look at the difference
in the gradient rather than in the voxel values.
As such, we define
the \emph{cycle difference image} as
\[
||(\partial_1(G_{P\to C}(G_{C\to P}(x))),\partial_2(G_{P\to C}(G_{C\to P}(x))))||_2
-||(\partial_1(x),\partial_2(x))||_2,
\]
where $\partial_1, \partial_2$ are Sobel's gradient operators \cite{Sobel}.
If both generators $G_{C\to P}$ and $G_{P\to C}$ work perfectly, 
the cycle difference image should constantly be zero.
Thus, we can use this cycle difference image to check how and where 
the generators may have failed.
To demonstrate this, we selected a pair of successfully learned models 
$G_{C\to P}$ and $G_{P\to C}$ and
an input image $x$ containing metal seeds and a couch, which were not seen in the training data.
The cycle difference image clearly indicates anomalous objects 
that were not learned during training (Figure \ref{fig:cyclediff}).
%Notice that contrast of cyclically translated image is enhanced in such regions as gas and bone where PlanCT has high contrast.
%On the other hand, contrast is reduced in abnormal regions.
%This is because there is more ambiguity in CBCT than PlanCT images,
%and hence, the generator $G_{C\to P}$ learned the contrast better than 
%$G_{P\to C}$.
This result shows that we can use the cycle difference as a simple safeguard against failures both in training and inference.
Note, however, the cycle difference is not a perfect measure for detecting failures.
It is necessary but not sufficient that $\Loss_{cycleA}$ is very small when conversion $x\mapsto G_{C\to P}(x)$ is successful.

\vspace{5mm}

There are two possible directions in which our method can be generalized.
We believe that the proposed method works properly with CBCT and PlanCT image datasets acquired in other institutions.
In addition, the proposed method should improve the image quality of organs other than the pelvis.
These two directions will be pursued in future research.

%%%%%
\section{Conclusion}\label{sec:conclusion}

We developed a synthetic approach based on CycleGAN to produce SynPlanCT images from CBCT images. The proposed approach relies only on unpaired and unaligned CBCT and PlanCT images for training.
The image quality of the synthesized PlanCT images substantially improved compared to those of the original CBCT.
% in terms of voxel values, spatial uniformity, and artifact suppression compared to those of the original CBCT. 
The anatomical structures of the original CBCT were well preserved in SynPlanCT.
In order to demonstrate the robustness of our method, we compared five models trained with the same structure and hyper-parameters, and observed an acceptable fluctuation of less than 10 HU.
The proposed method may be applied directly to 3D CBCT images reconstructed from a commercial CBCT scanner with a high computational efficiency.
The proposed method may enable soft tissue details to be more easily visualized.
% and the use of CBCT images for patient-specific treatment and dose calculation in radiotherapy.

\section*{Acknowledgments} 
Kaji was partially supported by JST PRESTO, Grant Number JPMJPR16E3 Japan.
Nawa was partially supported by a JSPS Grant-in-Aid for Young Scientists (B), 17K15799.
Imae was partially supported by a JSPS Grant-in-Aid for Scientific Research (C), 18K07667.
Nakamoto was partially supported by a JSPS Grant-in-Aid for JSPS Fellows, 18J00599, and a Grant-in-Aid for Early-Career Scientists, 18K15625.
Ohta was partially supported by a JSPS Grant-in-Aid for Early-Career Scientists, 18K15583.

We would like to thank Editage (\url{www.editage.jp}) for English language editing.

\section*{Conflict of Interest Statement}
This study was partially funded by Canon Medical Systems Corporation.

%\section*{Appendix}

%\section*{References}
%\bibliographystyle{amsplain}
%\bibliographystyle{iopart-num}
%\bibliographystyle{jphysicsB}

\bibliographystyle{medphy}  
\bibliography{ref.bib}